\documentclass[lettersize,journal]{IEEEtran}
\usepackage{amsmath,amsfonts}
\usepackage{algorithmic}
\usepackage{algorithm}
\usepackage{array}
\usepackage[caption=false,font=normalsize,labelfont=sf,textfont=sf]{subfig}
\usepackage{textcomp}
\usepackage{tabularx}
\usepackage{stfloats}
\usepackage{url}
\usepackage{verbatim}
\usepackage{graphicx}
\usepackage{cite}
\usepackage{amssymb}
\usepackage{booktabs}
\usepackage{multirow}
\usepackage[hidelinks]{hyperref}
\begin{document}

\title{GSO-Net: Visual State Machines for Hazardous Freight Transfer Compliance at Petrochemical Logistics Nodes}

\author{Yu~Xie$^{1}$,
        Bangshu~Xiong$^{1*}$,
        Zhibo~Rao$^{1}$,
        Rui~Gan$^{2}$,
        Chongxuan~Liu$^{2}$,
        and Zechu~Ouyang$^{2}$%
\thanks{$^{1}$Yu Xie, Bangshu Xiong, and Zhibo Rao are with the School of Information Engineering, Nanchang Hangkong University, Nanchang 330063, China (e-mail: xieyuharrison@163.com; xiongbs@126.com; raoxi36@foxmail.com).}%
\thanks{$^{2}$Rui Gan, Chongxuan Liu, and Zechu Ouyang are with Jiangxi Expressway Petrochemical Co., Ltd., Nanchang 330063, China (e-mail: 2362762050@qq.com; 1040345268@qq.com; oyzcjxja@126.com).}%
\thanks{$^{*}$Corresponding author: Bangshu Xiong.}}

\markboth{IEEE Transactions on Intelligent Transportation Systems}%
{Xie \MakeLowercase{\textit{et al.}}: GSO-Net: Visual State Machines for Hazardous Freight Transfer Compliance at Petrochemical Logistics Nodes}

\maketitle

\begin{abstract}
Hazardous-freight operations at petrochemical logistics nodes are safety-critical for intelligent transportation systems, yet existing vision benchmarks rarely address procedural compliance under realistic deployment constraints. In large infrastructure networks, cameras often operate under sparse round-robin polling, so transfer status must be inferred from incomplete observations and localized evidence. We present GSO-Net, a large-scale benchmark for visual understanding of standard operating procedures (SOPs) in petrochemical unloading scenarios. To our knowledge, GSO-Net is the first public benchmark dataset dedicated to visual SOP understanding in petrochemical hazardous-freight transfer scenarios. It contains over 50,000 independently sampled frames from 64 real expressway petrochemical logistics nodes and adopts an SOP-derived hierarchy linking 9 macroscopic procedural steps with 15 microscopic operational states. Two tasks are defined: joint detection of microscopic states and macroscopic steps as the core benchmark, and frame-level step classification as a diagnostic reference. Experiments with lightweight, transformer-based, open-vocabulary, and holistic models reveal a clear gap between object perception and transfer-stage understanding. Current models remain weak on contact-level state grounding, transient step recognition, and stage consistency, especially under sparse polling, tiny critical targets, and long-tailed operational evidence. GSO-Net provides a practical benchmark for fine-grained state perception and vision-based safety monitoring in hazardous freight transportation. The dataset is publicly available at \url{https://github.com/yuxieHarrison/GSO-Net}.
\end{abstract}


\section{Introduction}
\label{sec:introduction}

\IEEEPARstart{T}{he} safe transportation of hazardous freight is fundamental to transportation safety. Along expressway corridors, oil transport and unloading are essential to highway service support and petrochemical logistics. In such scenarios, tanker vehicles must follow a regulated SOP covering grounding, fire protection, hose connection, supervision, and disengagement. Any omission or violation may cause serious risks such as leakage, fire, or explosion. Automated SOP monitoring is therefore important for hazardous freight transportation safety.

Practical deployment differs from standard vision settings. Large infrastructure networks often monitor fixed camera feeds simultaneously, so they rely on \textit{round-robin visual polling}. In practice, each logistics node is typically observed for only 3--5\,s before switching to the next logistics node. The key challenge is not dense video understanding. It is reliable transfer-stage inference from sparse visual evidence.

This setting pushes the task beyond conventional object detection. The goal is not only to recognize visible objects. It is to determine procedural status from small visual cues across a hazardous freight-transfer scene in real deployments. Existing benchmarks do not fully capture this requirement. Most anomaly datasets, such as MVTec AD \cite{mvtec} and VisA \cite{visa}, focus on appearance defects. Safety and transportation datasets, such as PPE benchmarks \cite{ppe_construction} and traffic-scene datasets \cite{ref_trafficcam,ref_paralleleye}, verify the presence of objects or scene elements, but not whether a hazardous transfer procedure is being executed in the correct stage of a logistics node.

\begin{figure}[!t]
\centering
\includegraphics[width=1.02\columnwidth]{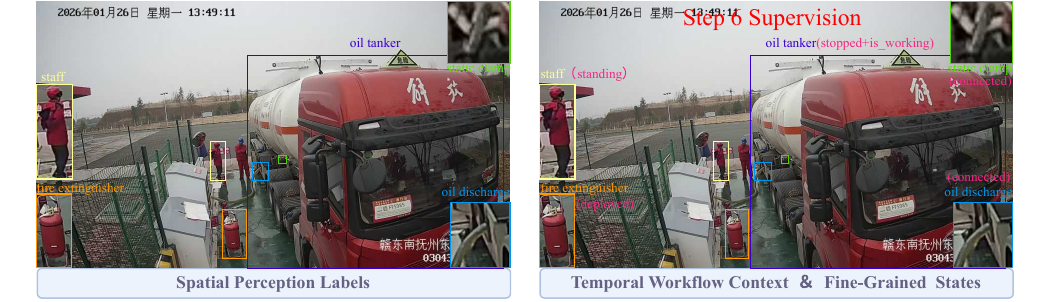}
\caption{Comparison of annotation paradigms. \textbf{Left (existing paradigm):} Traditional object detection identifies generic object categories but does not explicitly represent operational stages or states. \textbf{Right (GSO-Net paradigm):} GSO-Net organizes macroscopic procedural tags ($S_{macro}$) together with microscopic interactive states ($E_{micro}$), enabling state-aware analysis of hazardous freight-transfer scenarios at petrochemical logistics nodes.}
\label{fig:annotation_compare}
\end{figure}

As illustrated in Fig.~\ref{fig:annotation_compare}, object-centric perception is not sufficient for SOP-aware state understanding. Infrastructure-side compliance analysis must infer operational progress and procedural correctness from distributed visual evidence. This shift introduces three structural challenges:
\begin{itemize}
    \item \textbf{Spatiotemporal aliasing:} Adjacent workflow steps often share nearly identical global layouts, and their differences are confined to subtle local changes.
    \item \textbf{Extreme foreground-background imbalance:} Critical contact states may occupy only a few pixels in large scenes and are easily degraded by occlusion, illumination variation, and viewpoint shift.
    \item \textbf{Weak procedural priors in generic pre-training:} Foundation models provide strong semantic transfer across domains, but they do not explicitly encode hazardous-freight procedures or transfer-stage logic in practice.
\end{itemize}

These challenges suggest that infrastructure-side monitoring of hazardous freight transfer should be studied as a hierarchical state-understanding problem rather than as a pure detection task. A practical system must detect local entities, infer their functional states, and relate these cues to a broader transfer stage under sparse and intermittent observations.

To this end, we introduce \textbf{GSO-Net}, a benchmark for visual SOP understanding at petrochemical logistics nodes under sparse polling in real deployments. GSO-Net translates SOP rules into structured visual annotations by directly linking macroscopic procedural steps with microscopic operational states. It targets an underexplored setting where hazardous freight-transfer understanding depends on joint modeling of workflow order and localized functional states.

The main contributions are as follows:
\begin{enumerate}
    \item To our knowledge, we present \textbf{GSO-Net}, \textbf{the first publicly available dataset and benchmark} for visual SOP understanding in petrochemical hazardous freight-transfer scenarios. It contains more than 50,000 independently sampled frames from 64 unconstrained logistics nodes in real-world infrastructure settings.
    \item We propose a \textbf{hierarchical annotation framework} coupling 9 macroscopic procedural steps with 15 microscopic operational states. It translates SOP descriptions into explicit visual evidence. This enables stage-aware understanding in petrochemical transfer scenarios.
    \item We establish a \textbf{benchmark protocol}. It centers on joint step-state detection. It is complemented by frame-level step classification. We also provide lightweight, transformer-based, open-vocabulary, and holistic baselines for \textbf{systematic and comprehensive} evaluation.
    \item We provide \textbf{empirical analysis}. It reveals the limitations of current architectures. These limitations arise under sparse polling, tiny safety-critical cues, and long-tailed operational evidence. Our analysis thereby highlights the practical challenges of visual SOP understanding in petrochemical logistics nodes.
\end{enumerate}

\section{Related Work}
\label{sec:related}

GSO-Net targets visual SOP understanding for hazardous freight transfer at petrochemical logistics nodes under sparse infrastructure-side polling. It builds on related efforts in safety monitoring, procedural understanding, and fine-grained visual perception, while aiming to fill a gap at their intersection: explicit workflow order, localized operational states, and SOP-oriented supervision under sparse observations. To make these differences explicit, Table~\ref{tab:dataset_comparison} compares representative benchmarks with GSO-Net along these three dimensions.

\begin{table*}[!t]
\caption{Comparison of GSO-Net with representative related benchmarks. Existing datasets typically emphasize only one aspect of the problem, such as anomaly cues, workflow order, or safety-object perception. By contrast, GSO-Net jointly models workflow order and localized operational states for hazardous freight-transfer SOP understanding at petrochemical logistics nodes under sparse visual polling.}
\label{tab:dataset_comparison}
\centering
\footnotesize
\renewcommand{\arraystretch}{1.12}
\setlength{\tabcolsep}{4.2pt}
\begin{tabularx}{\textwidth}{
>{\raggedright\arraybackslash}p{2.35cm}
>{\centering\arraybackslash}p{1.32cm}
>{\raggedright\arraybackslash}p{2.70cm}
>{\raggedright\arraybackslash}p{2.20cm}
>{\raggedright\arraybackslash}X
>{\centering\arraybackslash}p{0.88cm}
>{\centering\arraybackslash}p{0.88cm}
>{\centering\arraybackslash}p{0.88cm}
}
\toprule
\textbf{Dataset} & \textbf{Venue} & \textbf{Focus} & \textbf{Scale} & \textbf{Supervision} & \textbf{Order} & \textbf{States} & \textbf{SOP} \\
\midrule

ParallelEye \cite{ref_paralleleye}
& T-ITS'19
& Virtual traffic-scene perception
& 40,251 images
& Multi-task dense annotations
& $\times$ & $\times$ & $\times$ \\

Assembly101 \cite{assembly101}
& CVPR'22
& Procedural activity understanding
& 4,321 videos / $>$1M segments
& Dense temporal action annotations
& $\checkmark$ & $\times$ & $\times$ \\

VisA \cite{visa}
& ECCV'22
& Multi-instance anomaly detection
& 10,821 images
& Image-/pixel-level anomaly labels
& $\times$ & $\times$ & $\times$ \\

MVTec LOCO AD \cite{loco}
& IJCV'22
& Structural and logical anomaly detection
& 3,644 images
& Image-/pixel-level anomaly labels
& $\times$ & $\times$ & $\times$ \\

MVTec 3D-AD \cite{mvtec3d}
& VISAPP'22
& 3D anomaly detection
& 4,147 depth maps / point clouds
& Pixel-precise anomaly annotations
& $\times$ & $\times$ & $\times$ \\

Eyecandies \cite{eyecandies}
& ACCV'22
& Multimodal anomaly detection
& 10 categories / RGB-D
& Image-/pixel-level anomaly labels
& $\times$ & $\times$ & $\times$ \\

VoxelScape \cite{ref_voxelscape}
& T-ITS'23
& Simulated 3D traffic perception
& 100K point-cloud scans
& Point-wise labels + 3D boxes
& $\times$ & $\times$ & $\times$ \\

OpenPack \cite{openpack}
& PerCom'24
& Industrial workflow recognition
& 53.0 hours / 20,129 labels
& Multimodal operation annotations
& $\checkmark$ & $\times$ & $\times$ \\

Real-IAD \cite{realiad}
& CVPR'24
& Real-world multi-view anomaly detection
& 150K images / 30 objects
& Image-/pixel-level anomaly labels
& $\times$ & $\times$ & $\times$ \\

MMAD \cite{mmad}
& ICLR'25
& Industrial multimodal reasoning
& 8,366 images / 39,672 QAs
& QA-style multimodal annotations
& $\times$ & $\times$ & $\times$ \\

TrafficCAM \cite{ref_trafficcam}
& T-ITS'25
& Fixed-camera traffic flow segmentation
& 4,364 labeled + 58,689 unlabeled frames
& Pixel-/instance-level annotations
& $\times$ & $\times$ & $\times$ \\

MVTec AD 2 \cite{mvtecad2}
& IJCV'26
& Advanced anomaly detection
& 8 scenarios / 8,000+ images
& Image-/pixel-level anomaly labels
& $\times$ & $\times$ & $\times$ \\

\midrule
\textbf{GSO-Net (Ours)}
& \textbf{2026}
& \textbf{Hazardous freight-transfer SOP understanding}
& \textbf{50,325 frames / 321,432 boxes}
& \textbf{Macro-step labels + micro-state boxes}
& $\checkmark$ & $\checkmark$ & $\checkmark$ \\
\bottomrule
\end{tabularx}

\vspace{0.35em}
\footnotesize
\textit{Note:} ``Order'' indicates whether workflow order is an explicit target. ``States'' indicates whether localized functional states are explicitly annotated or evaluated. ``SOP'' indicates whether the benchmark is explicitly designed for hazardous procedural understanding.
\end{table*}

\subsection{Benchmarks for Safety Monitoring and Procedural Compliance}

Existing public benchmarks for safety monitoring usually focus on abnormality, object presence, or violation detection rather than SOP understanding. In industrial inspection, datasets such as VisA \cite{visa}, MVTec LOCO AD \cite{loco}, MVTec 3D-AD \cite{mvtec3d}, Eyecandies \cite{eyecandies}, Real-IAD \cite{realiad}, Real3D-AD \cite{real3dad}, Real-IAD D$^3$ \cite{realiad_d3}, M3-AD \cite{m3ad}, and MVTec AD 2 \cite{mvtecad2} mainly study whether an object, part, or scene is defective or anomalous. These datasets are valuable for inspection-oriented perception, but their supervision centers on defectiveness rather than operational correctness. A frame may be visually normal in such benchmarks yet still correspond to an incorrect stage in a hazardous freight-transfer procedure.

Safety-oriented datasets are closer to our application setting, but they still do not capture the supervision structure required in our problem. PPE benchmarks \cite{ppe_construction} focus on worker protection and equipment presence, while transportation-side benchmarks such as TrafficCAM \cite{ref_trafficcam}, ParallelEye \cite{ref_paralleleye}, and VoxelScape \cite{ref_voxelscape} cover fixed-camera traffic flow segmentation, synthetic traffic-scene generation, and simulated 3D traffic perception. These datasets are valuable for traffic-scene perception and infrastructure-side monitoring, but they still focus on traffic participants, scene elements, or general perception targets rather than node-level procedural compliance. In these settings, the key question is typically whether a target object or scene element is present. In contrast, the critical question in hazardous freight transfer is whether grounding, extinguisher deployment, hose connection, personnel supervision, and disengagement occur at the correct procedural stage.

Therefore, the main gap in existing benchmarks is not simply scene domain, but supervision structure. Prior datasets do not provide a public benchmark that jointly models workflow order, localized operational states, and SOP-oriented compliance in petrochemical hazardous freight-transfer scenarios. GSO-Net addresses this gap by explicitly linking macroscopic procedural stages with microscopic state annotations in a real infrastructure-side transportation setting.

\subsection{Procedural Understanding Under Sparse Infrastructure Observation}

Procedural understanding has been extensively studied in video understanding and workflow analysis. Benchmarks such as Assembly101 \cite{assembly101}, EPIC-KITCHENS-100 \cite{epic100}, Ego4D \cite{ego4d}, HD-EPIC \cite{hdepic}, MLVU \cite{mlvu}, and OpenPack \cite{openpack} provide rich supervision for action recognition, anticipation, temporal parsing, and long-horizon activity understanding. These works have substantially advanced procedural reasoning, especially in environments where temporal continuity is strong and activities can be observed through dense sequential evidence.

However, this assumption does not hold in our deployment setting. In large infrastructure networks, many fixed cameras must share limited computational and transmission resources, so monitoring systems often rely on round-robin polling rather than continuous observation. Under this regime, each logistics node is observed only intermittently, and critical procedural transitions may appear in only one or two frames. As a result, stage recognition cannot depend on dense motion cues, temporal tracking, or long contiguous clips in the way that many existing procedural benchmarks assume.

This distinction is fundamental. GSO-Net is not designed for dense activity parsing from continuous video, but for procedural inference from sparse infrastructure observations. The model must recover workflow stage from incomplete visual evidence and small operational cues, rather than from stable temporal evolution. Existing procedural benchmarks therefore provide important conceptual context, but do not capture the sparse, discontinuous observation regime that characterizes infrastructure-side hazardous freight monitoring.

\subsection{Fine-Grained State for Stage-Aware SOP Analysis}

A second challenge of our setting is that procedural understanding depends on fine-grained operational evidence. Modern detection methods, including Faster R-CNN \cite{fasterrcnn}, DETR \cite{detr}, DINO \cite{dino}, RT-DETR \cite{ref_rtdetr}, Relation DETR \cite{ref_relationdetr}, Hyper-YOLO \cite{ref_hyperyolo}, Mr. DETR \cite{ref_mrdetr}, and RF-DETR \cite{ref_rfdetr}, have substantially improved localization and representation learning. Transportation studies such as CLPD \cite{ref_clpd}, SGV3D \cite{ref_sgv3d}, VoxelScape \cite{ref_voxelscape}, and infrastructure-side point cloud detection research \cite{ref_mamfnet} further show how perception systems must adapt to realistic deployment conditions, while tiny-object benchmarks such as CST Anti-UAV \cite{antiuav} and recent T-ITS analysis of image- and video-based small object detection \cite{ref_sodguide25} highlight the fragility of small-target recognition under severe scale compression. These works are highly relevant because decisive visual evidence in petrochemical unloading scenes, such as clamp status or hose connection status, is often tiny, localized, and easily degraded.

Recent foundation and open-vocabulary models further strengthen semantic transfer. CLIP \cite{clip}, GLIP \cite{glip}, Grounding DINO \cite{ref_groundingdino}, YOLO-World \cite{ref_yoloworld}, LLMDet \cite{ref_llmdet}, Dynamic-DINO \cite{ref_dynamicdino}, and OW-OVD \cite{ref_owovd} improve phrase-level grounding and category generalization, while DINOv2 \cite{ref_dinov2}, Qwen2.5-VL \cite{ref_qwen25vl}, InternVL3 \cite{ref_internvl3}, AnomalyGPT \cite{anomalygpt}, and MMAD \cite{mmad} provide stronger visual or multimodal priors for inspection-oriented reasoning. Part-aware annotations in PACO \cite{paco}, together with fine-grained traffic-scene benchmarks such as TrafficCAM \cite{ref_trafficcam} and ParallelEye \cite{ref_paralleleye}, also point toward more detailed visual perception.

However, stronger perception alone is insufficient for GSO-Net. Object recognition does not directly resolve whether local operational states are functionally valid or procedurally consistent with the current workflow stage. In our setting, microscopic state grounding and macroscopic stage inference are tightly coupled. The core challenge is to bridge localized semantic evidence and global procedural awareness, enabling a transition from microscopic cue understanding to macroscopic state cognition. Existing perception models improve localization and semantic transfer, but they do not explicitly evaluate or supervise this relationship under sparse polling. GSO-Net is designed to make this gap measurable through unified annotations and evaluation.

\section{GSO-Net Construction and Annotation Protocol}
\label{sec:dataset}

\begin{figure*}[!t]
\centering
\includegraphics[width=1.0\textwidth]{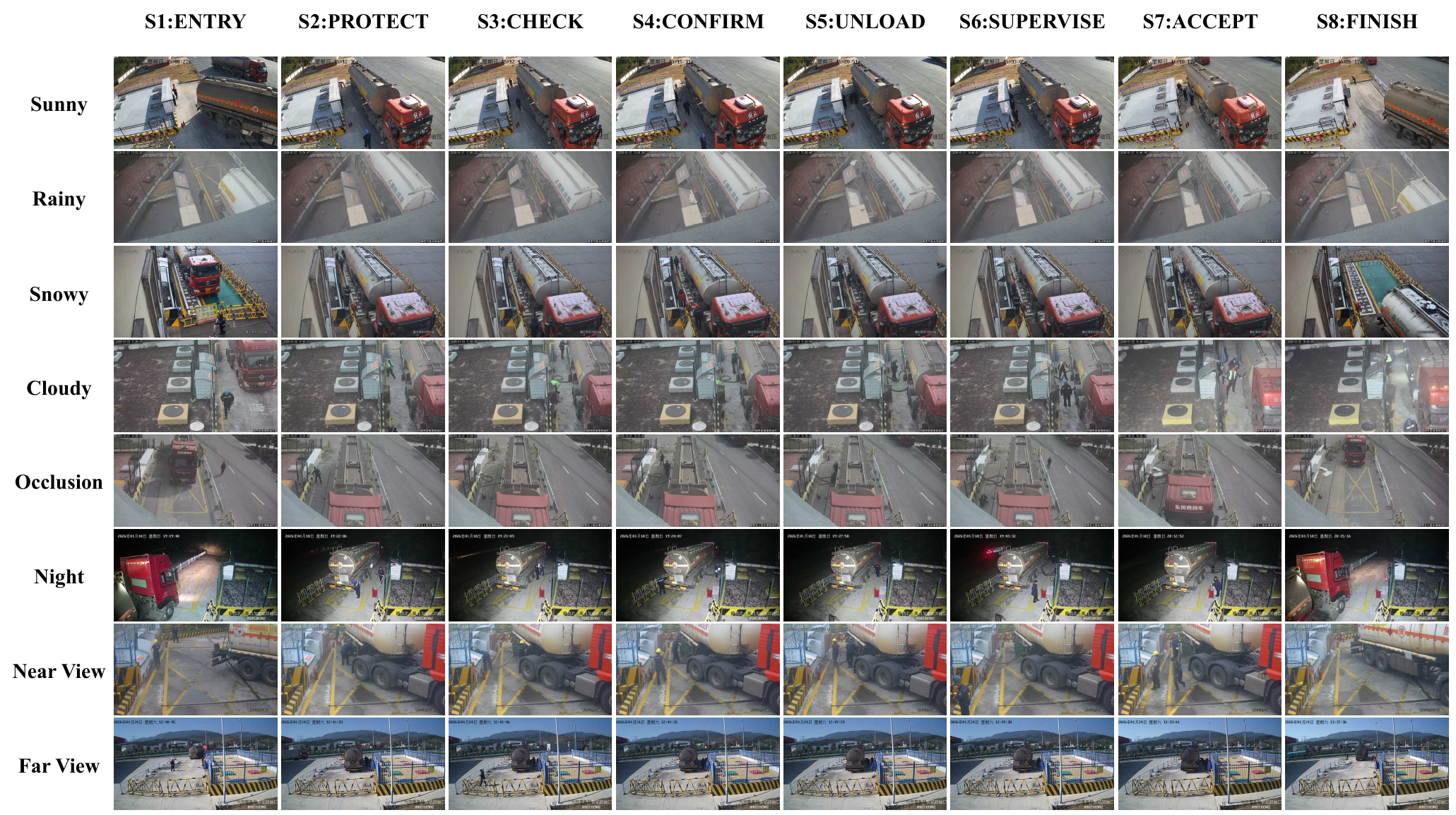}
\vspace{-1.5em}
\caption{Overview of the GSO-Net benchmark. The dataset matrix summarizes the diversity of GSO-Net across illumination conditions (day and night), weather conditions (rain and snow), dynamic occlusions, and the chronological progression of the 9 macroscopic SOP steps in petrochemical freight-transfer operations.}
\label{fig:dataset_overview}
\end{figure*}

This section describes the construction of GSO-Net. Our goal is to translate SOP rules for petrochemical freight-transfer operations into a computable visual benchmark. As shown in Fig.~\ref{fig:dataset_overview}, GSO-Net covers substantial diversity in stage, weather, illumination, viewpoint, and occlusion.

\subsection{SOP Taxonomy and Hierarchical Semantic Mapping}
Industrial SOPs are designed to reduce operational risk under strict physical and procedural constraints. We translate these textual procedures into a hierarchical taxonomy, shown in Table~\ref{tab:taxonomy_table}. This taxonomy links continuous workflow phases to explicit and interpretable visual evidence.

\textbf{Level 1: Macroscopic procedural steps ($S_{macro}$).}  
We partition the unloading workflow into nine discrete and chronologically dependent phases ($S_0$ to $S_8$). Each phase corresponds to a distinct operational stage and a distinct safety requirement.
\begin{itemize}
    \item \textit{$S_0$: Idle.} The logistics node is in an idle state, and no unloading operation is active.
    \item \textit{$S_1$: Entry.} The oil tanker arrives and enters the operating zone.
    \item \textit{$S_2$: Safety protection.} Grounding and fire-protection measures are deployed before unloading begins.
    \item \textit{$S_3$: Safety check.} Personnel verify that the deployed safety measures are correctly positioned and accessible.
    \item \textit{$S_4$: Pre-unload confirm.} The unloading hose is positioned near the discharge port, but fuel transfer has not started.
    \item \textit{$S_5$: Oil unloading.} The unloading hose is physically connected to the tanker port, and fuel transfer is in progress.
    \item \textit{$S_6$: Supervision.} Fuel transfer is underway, and personnel remain on site for supervision.
    \item \textit{$S_7$: Receipt acceptance.} Fuel transfer completion is verified, and the hose is disconnected and drained.
    \item \textit{$S_8$: Post-processing.} The oil tanker departs, and the logistics node returns to the idle state.
\end{itemize}

\textbf{Level 2: Microscopic operational states ($E_{micro}$).}  
We ground the procedural description into bounding-box annotations and define 15 fine-grained operational states, such as \textit{clamp\_connected} and \textit{discharge\_disconnected}. These states provide visual evidence for stage understanding.

\begin{table*}[!t]
\caption{Hierarchical taxonomy and semantic definitions of microscopic states ($\mathcal{E}_{micro}$) in GSO-Net.}
\label{tab:taxonomy_table}
\centering
\footnotesize
\renewcommand{\arraystretch}{1.22}
\setlength{\tabcolsep}{8pt}
\begin{tabular}{p{2.0cm} p{3.8cm}@{\hspace{2pt}}p{11.1cm}}
\toprule
\textbf{Entity Category} & \textbf{Micro-State Label} & \textbf{Semantic Operational Significance} \\
\midrule
\textbf{Oil Tanker} & 00: tanker\_moving\_true & \textbf{High-risk anomaly:} Unauthorized vehicle displacement during active fluid transfer, with elevated risk of structural failure or leakage. \\
& 01: tanker\_moving\_false & \textbf{Standard transit:} Vehicle in motion for entry, exit, or positioning, without active discharge. \\
& 02: tanker\_stopped\_true & \textbf{Active transfer:} Stationary vehicle during ongoing fuel discharge. \\
& 03: tanker\_stopped\_false & \textbf{Idle/standby:} Stationary vehicle without active discharge. \\
\midrule
\textbf{Personnel} & 04: staff\_standing & \textbf{Supervision:} On-site personnel monitoring the operational state. \\
& 05: staff\_bending & \textbf{Auxiliary inspection:} Personnel adjusting or checking components outside the discharge action. \\
& 06: staff\_walking & \textbf{Task transit:} Personnel moving toward a specific operational or safety-related task. \\
& 07: staff\_operating & \textbf{Active interaction:} Direct manual engagement with mechanical systems. \\
\midrule
\textbf{Static Clamp} & 08: clamp\_connected & \textbf{Validated grounding:} Physical grounding is visibly established. \\
& 09: clamp\_disconnected & \textbf{Safety violation:} Grounding is absent during hazardous operation. \\
& 10: clamp\_floating & \textbf{Grounding procedure:} Personnel are deploying or adjusting the static clamp. \\
\midrule
\textbf{Extinguisher} & 11: extinguisher\_deployed & \textbf{Safety readiness:} Fire suppression equipment is positioned at the required safety point. \\
& 12: extinguisher\_carried & \textbf{Manual transit:} Fire suppression equipment is being moved by personnel. \\
\midrule
\textbf{Oil Discharge} & 13: discharge\_connected & \textbf{Secured interface:} The discharge hose is mechanically connected at the port. \\
& 14: discharge\_disconnected & \textbf{Disengaged interface:} The discharge hose is visibly disconnected from the port. \\
\bottomrule
\end{tabular}
\end{table*}

\subsection{Step-State Correspondence Derived from SOPs}
A central design principle of GSO-Net is the correspondence between macroscopic steps and microscopic states. This correspondence is directly derived from SOP documents and operational practice. It is used to organize the annotation system and interpret stage-specific visual evidence.

At the macroscopic level, each frame is assigned one of the 9 procedural steps. At the microscopic level, the frame is annotated with visible state-specific objects and interactions. The two levels are related by operational meaning rather than by a formally defined rule engine. In other words, we do not define a complete logical protocol, symbolic solver, or explicit constraint-based evaluation metric. Instead, we use SOPs to guide the hierarchical label design and the semantic interpretation of the full dataset.

For example, the \textit{Continuous Supervision} stage is typically associated with states such as \textit{clamp\_connected}, \textit{discharge\_connected}, and the visible presence of on-site personnel. Similarly, pre-unloading and post-processing stages are characterized by different local state combinations and different operational focus. These correspondences help explain why stage recognition in GSO-Net depends on small and localized evidence rather than on global scene layout alone.

This design choice is important. It keeps the dataset faithful to what is actually annotated. GSO-Net provides a structured benchmark for hierarchical SOP understanding. It does not claim to implement a full logic-verification protocol. Instead, it offers a benchmark foundation on which future work may build more explicit reasoning or verification modules.

\subsection{Independent Frame Sampling and Environmental Complexity}
To reflect practical deployment, data collection was conducted at scale. We sourced surveillance feeds from 64 active expressway petrochemical logistics nodes over eight months.

We did not extract continuous high-frame-rate clips. Instead, we adopted an independent frame-sampling strategy. This design matches the round-robin visual polling widely used in practical infrastructure-side edge systems.

These logistics nodes are not controlled laboratory environments. They are unconstrained real-world logistics nodes. The dataset therefore includes strong diurnal changes, seasonal variation, and adverse weather. A substantial portion of the dataset comes from night scenes and non-ideal weather conditions. From the full surveillance pool, we curated more than 50,000 high-fidelity frames. This design reduces site-specific bias and discourages simple background memorization.

\subsection{Annotation Procedure and Quality Assurance}
The annotation process was carried out over four months by the core team of the smart logistics node project. All labeling rules were derived from operational procedures used in frontline petrochemical unloading. To improve annotation reliability, a three-stage quality-assurance protocol was adopted.

\textbf{Stage 1: State-aware bounding-box annotation.}  
All critical entities and operational states were manually annotated at the bounding-box level. During this stage, \textit{functional engagement} was explicitly distinguished from \textit{spatial proximity}. For example, a grounding clamp was labeled as ``connected'' only when the visible evidence supported an actual mechanical connection rather than mere spatial closeness to the tanker.

\textbf{Stage 2: Frontline expert verification.}  
Because several microscopic states are visually subtle and operationally sensitive, targeted verification was performed with experienced gas station operators. Ambiguous samples were reviewed against real operational practice to ensure that the assigned labels remained consistent with actual procedural meaning.

\textbf{Stage 3: Multi-round semantic and procedural review.}  
After the initial annotation and expert verification, the labels were further reviewed for semantic consistency and procedural interpretability. Special attention was given to whether microscopic states were supported by visible evidence and whether macroscopic step labels were compatible with the observed operational context. Importantly, this review did not enforce that every sample satisfy an ideal SOP-compliant state. Samples showing unsafe, anomalous, or non-compliant operations were intentionally retained when such conditions were clearly visible. Samples with uncertain labels, inconsistent criteria, or difficult visual evidence were returned for group discussion and manual revision until a consensus label was reached.

This protocol ensures that the benchmark faithfully reflects visible operational evidence, real procedural context, and practically relevant abnormal cases rather than only idealized SOP-compliant operational settings.

\subsection{Dataset Characteristics and Structural Difficulty}
Statistical analysis further shows that GSO-Net is inherently challenging in several dimensions.

\begin{figure*}[!t]
\captionsetup[subfloat]{font=small}
\centering
\begin{minipage}[b]{0.51\textwidth}
    \centering
    \subfloat[\textbf{Macroscopic Spatiotemporal Density}]{%
        \raisebox{0.7cm}{\includegraphics[width=\linewidth]{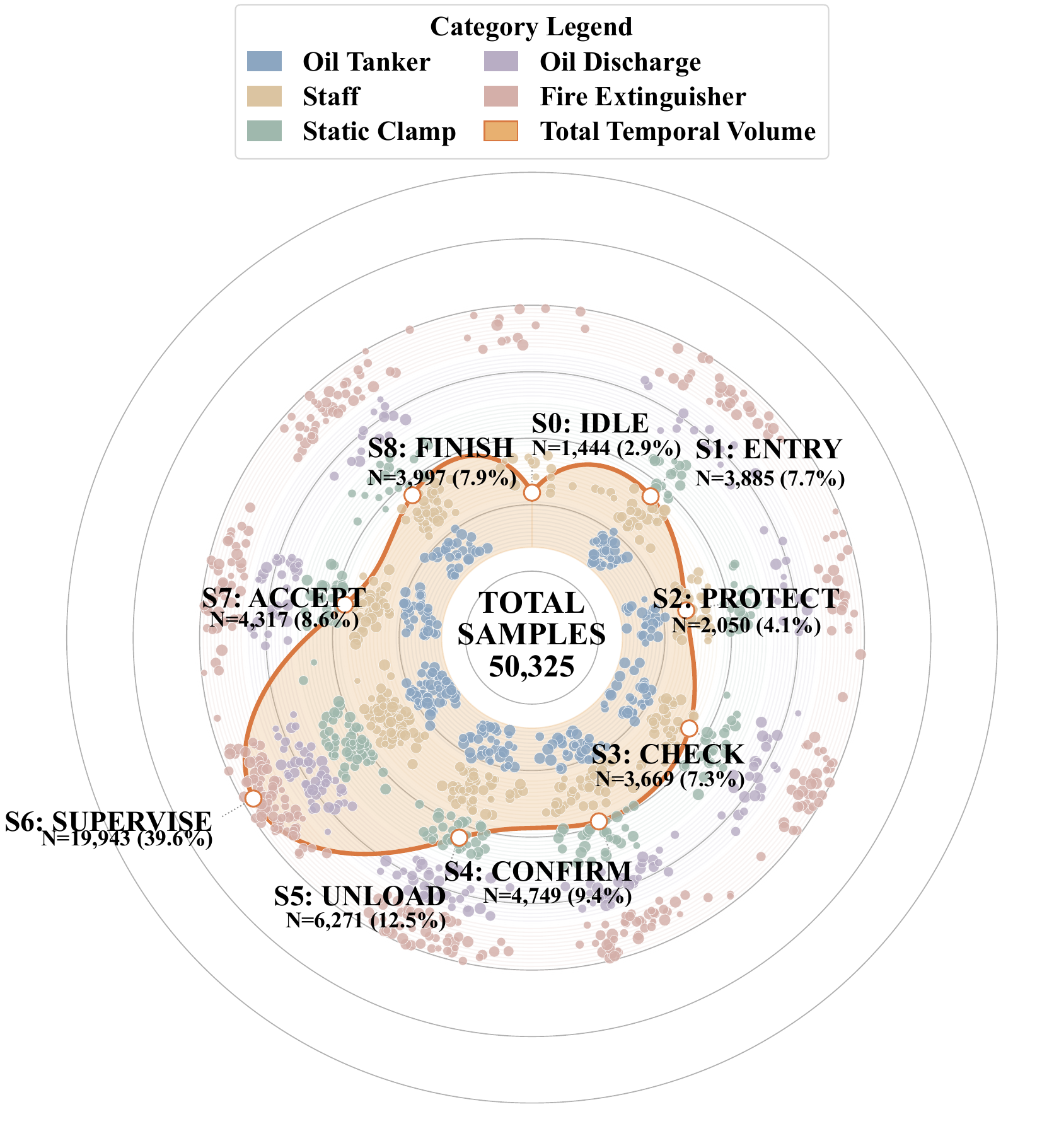}}%
        \label{fig:char_density}%
    }
\end{minipage}\hfill
\begin{minipage}[b]{0.48\textwidth}
    \centering
    \subfloat[\textbf{Step-State Co-occurrence}]{\includegraphics[width=\linewidth]{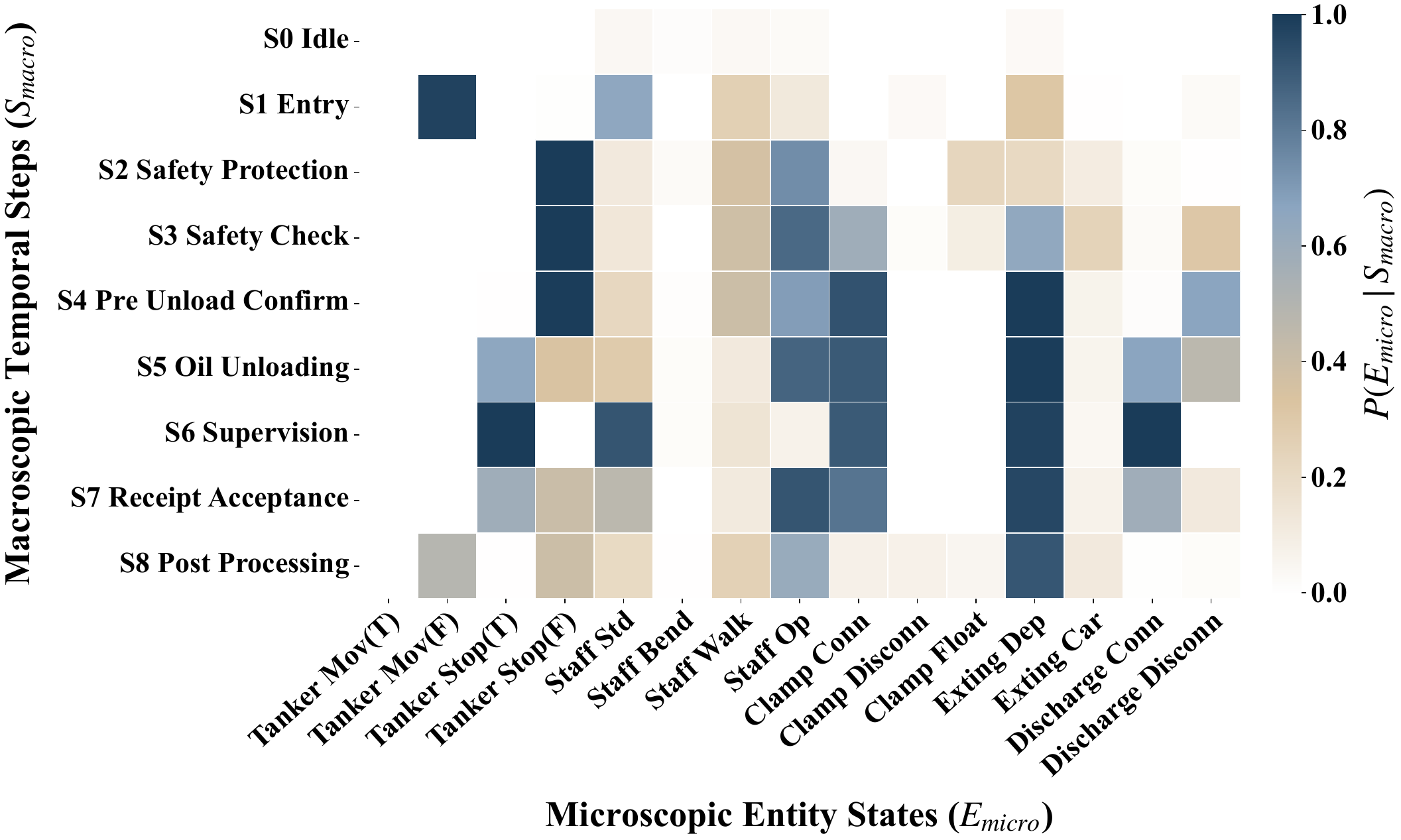}\label{fig:char_heatmap}}\\
    \vspace{1.5em}
    \subfloat[\textbf{Temporal Focus Shift}]{\includegraphics[width=\linewidth]{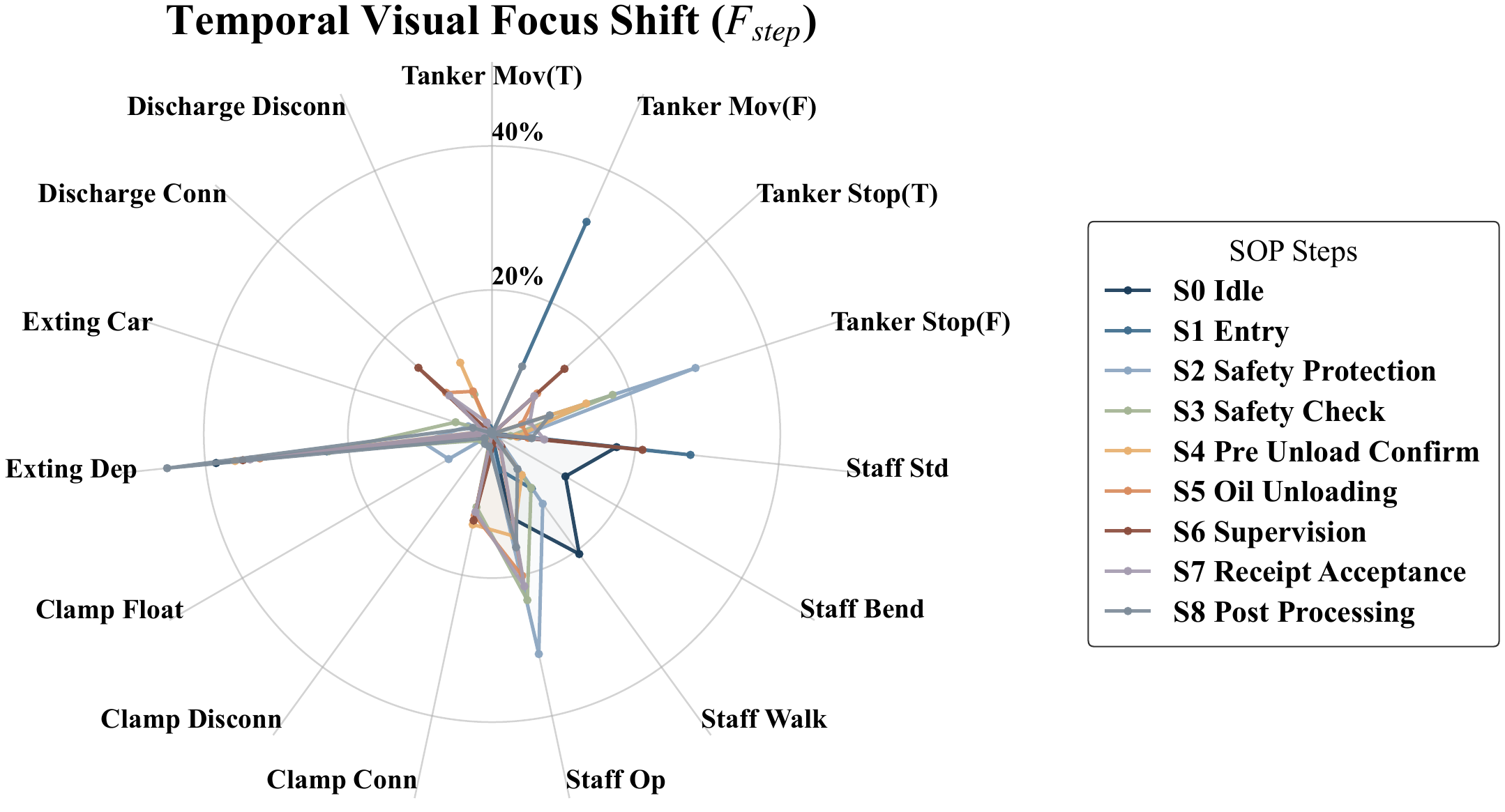}\label{fig:char_radar}}
\end{minipage}
\caption{Macroscopic and state-aware dynamics of GSO-Net. \textbf{(a)} Spatiotemporal density across the operational lifecycle. \textbf{(b)} Conditional probability heatmap $P(E_{micro}\mid S_{macro})$, showing the correspondence between macro steps and micro states. \textbf{(c)} Temporal focus fingerprints, showing how semantic focus shifts across adjacent SOP phases.}
\label{fig:macro_logic_dynamics}
\end{figure*}

As summarized in Fig.~\ref{fig:macro_logic_dynamics}, GSO-Net exhibits coupled macroscopic and microscopic structure across the SOP lifecycle.

\textbf{Operational lifecycle structure.}  
Fig.~\ref{fig:char_density} shows strong variation across the workflow timeline. Long-duration standby and supervision stages dominate the temporal distribution.

\begin{figure*}[!t]
\centering
\includegraphics[width=0.98\textwidth]{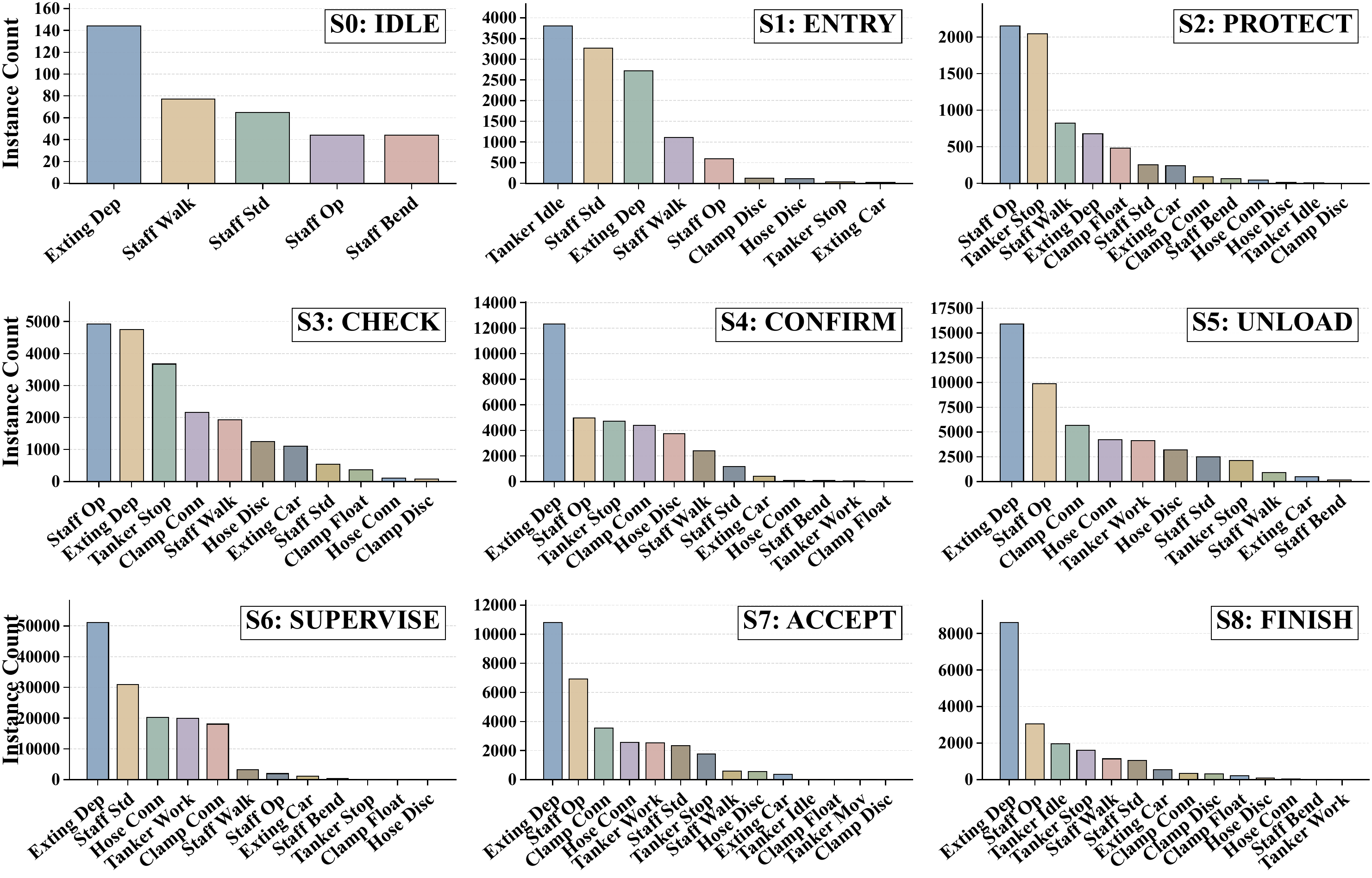}
\caption{Microscopic scale imbalance in GSO-Net. The entity distribution is strongly long-tailed across the 9 SOP stages. Static and frequent states dominate the benchmark, while safety-critical interaction states appear much less often. Step $S_6$ contributes the largest number of samples because it corresponds to the longest operational phase. This distribution makes the benchmark difficult for standard detectors, especially on rare but safety-critical categories.}
\label{fig:longtail_grid}
\end{figure*}

\textbf{Long-tailed microscopic distribution.}  
Fig.~\ref{fig:longtail_grid} shows that GSO-Net exhibits a pronounced long-tail distribution. Frequent states, such as \textit{tanker\_stopped\_true} and \textit{extinguisher\_deployed}, contain many instances, whereas safety-critical interaction states, such as \textit{clamp\_floating} and \textit{clamp\_disconnected}, are rarer. Step $S_6$ contributes the most samples because fuel transfer lasts longer than the preparatory and post-processing stages. This realistic imbalance also makes the benchmark challenging for generic detectors.

\textbf{Step-state correspondence.}  
Fig.~\ref{fig:char_heatmap} visualizes the empirical relationship between $S_{macro}$ and $E_{micro}$, showing that specific microscopic states cluster around specific procedural stages.

\textbf{Temporal focus shift.}  
Fig.~\ref{fig:char_radar} shows that adjacent SOP phases may share similar global views while differing in localized operational evidence. The model therefore needs dynamic semantic focus rather than global scene matching.

\subsection{Dataset Statistics}
\label{sec:dataset_stats}

We summarize the benchmark with statistics in Tables~\ref{tab:instance_counts} and \ref{tab:env_stats}. Table~\ref{tab:instance_counts} reports counts of all 15 microscopic states in the training and validation sets. The numbers confirm the long-tail nature of GSO-Net. Head classes such as \textit{staff\_standing} and \textit{extinguisher\_deployed} have tens of thousands of instances, whereas tail classes such as \textit{clamp\_floating} and \textit{staff\_bending} contain fewer than 1,100.

Table~\ref{tab:env_stats} summarizes the environmental and spatial diversity of GSO-Net. The benchmark contains 50,325 curated frames from 64 independent logistics nodes, with clear daylight accounting for only 42.2\% of the data. The remaining frames cover night scenes, adverse weather, viewpoint variation, and heavy occlusion, supporting evaluation under realistic degradation rather than controlled-scene pattern matching.

\begin{table}[!t]
\caption{Bounding-box instance distribution across the training and validation sets for the 15 microscopic states in GSO-Net.}
\label{tab:instance_counts}
\centering
\renewcommand{\arraystretch}{1.1}
\setlength{\tabcolsep}{8pt}
\begin{tabular}{lrrr}
\toprule
\textbf{Microscopic State ($E_{micro}$)} & \textbf{Train} & \textbf{Val} & \textbf{Total} \\
\midrule
tanker\_moving\_true         &       4 &       0 &       4 \\
tanker\_moving\_false        &   4,646 &   1,130 &   5,776 \\
tanker\_stopped\_true        &  21,736 &   4,899 &  26,635 \\
tanker\_stopped\_false       &  12,961 &   3,033 &  15,994 \\
staff\_standing              &  34,141 &   7,907 &  42,048 \\
staff\_bending               &     427 &     305 &     732 \\
staff\_walking               &   9,847 &   2,405 &  12,252 \\
staff\_operating             &  28,301 &   6,187 &  34,488 \\
clamp\_connected             &  27,668 &   6,616 &  34,284 \\
clamp\_disconnected          &     509 &       7 &     516 \\
clamp\_floating              &   1,004 &      92 &   1,096 \\
extinguisher\_deployed       &  88,430 &  18,641 & 107,071 \\
extinguisher\_carried        &   3,977 &     337 &   4,314 \\
discharge\_connected         &  22,353 &   4,931 &  27,284 \\
discharge\_disconnected      &   7,638 &   1,300 &   8,938 \\
\midrule
\textbf{Total Boxes}         & \textbf{263,642} & \textbf{57,790} & \textbf{321,432} \\
\bottomrule
\end{tabular}
\end{table}

\begin{table}[!t]
\caption{Distribution of frames across environmental conditions, viewpoints, and visual challenges in GSO-Net.}
\label{tab:env_stats}
\centering
\setlength{\tabcolsep}{8pt}
\begin{tabular}{lccc}
\toprule
\textbf{Condition / Viewpoint} & \textbf{\# Nodes} & \textbf{\# Images} & \textbf{Percentage} \\
\midrule
Daylight (Clear)         & 27 & 21,215 & 42.2\% \\
Night                    & 9  &  6,058 & 12.0\% \\
Near View                & 8  &  7,029 & 14.0\% \\
Far View                 & 7  &  6,085 & 12.1\% \\
Snow \& Cloud            & 5  &  3,426 &  6.8\% \\
Rain                     & 4  &  2,907 &  5.8\% \\
Occlusion                & 4  &  3,605 &  7.2\% \\
\midrule
\textbf{Total}           & \textbf{64} & \textbf{50,325} & \textbf{100.0\%} \\
\bottomrule
\end{tabular}
\end{table}

\subsection{Ethical Compliance and Data Integrity}
Data collection was conducted within the authorized scope of our smart logistics node project. All video feeds were acquired through standard workplace safety monitoring systems. To preserve operational fidelity, we retained original image frames during dataset construction. The acquisition, annotation, and release pipeline follows local industrial data-governance requirements and project-level ethical constraints.

\section{Experiments and Analysis}
\label{sec:experiments}

We evaluate GSO-Net as a benchmark rather than a task-specific method. The experiments test whether current models can recover procedural meaning from localized evidence under sparse visual polling at hazardous freight logistics nodes. Two tasks are considered. \textbf{Task 1} is the core benchmark and evaluates joint detection of microscopic states and macroscopic steps. \textbf{Task 2} is a diagnostic reference that evaluates frame-level classification of macroscopic SOP steps without explicit state grounding. Together, these tasks expose the effects of sparse polling, long-tailed states, small critical targets, and stage-sensitive visual evidence.

\subsection{Benchmark Protocol and Experimental Setup}
\label{sec:exp_setup}

A strict \textit{cross-node split} is adopted. Random frame-level splitting is not used because it introduces scene leakage and allows models to memorize node-specific backgrounds. The training set contains 40,976 images from 51 logistics nodes. The validation set contains 9,349 images from 13 unseen logistics nodes. This protocol measures generalization to new nodes rather than memorization within the same logistics node.

Two benchmark tasks are defined. The distinction is intentional: Task 1 evaluates the full hierarchical formulation of GSO-Net, whereas Task 2 quantifies the limitations of transfer-stage recognition when localized state evidence is not explicitly modeled in holistic prediction.

\textbf{Task 1: Joint detection.} Each detector predicts 24 categories in a unified label space, including 15 microscopic states and 9 macroscopic procedural steps. The microscopic states use their original bounding-box annotations, while the macroscopic steps are represented as full-frame targets. Thus, AP for microscopic states evaluates localized state detection, whereas AP for macroscopic steps evaluates image-level step recognition. Performance is measured by $\mathrm{AP}_{50}$, $\mathrm{AP}_{50\text{-}95}$, and scale-aware metrics $\mathrm{AP}_{S}$, $\mathrm{AP}_{M}$, and $\mathrm{AP}_{L}$. Since the unified label space contains both localized and full-frame targets, the scale-aware metrics should be interpreted with care. This is the main evaluation because GSO-Net focuses on stage understanding together with localized operational evidence under sparse infrastructure-side observation.

\textbf{Task 2: Frame-level step classification.} Each classifier assigns one of the 9 macroscopic SOP steps to each full image based on global visual evidence. Performance is evaluated using Top-1 Accuracy, Top-5 Accuracy, and Macro F1. This task is weaker than Task 1 and serves as a diagnostic reference for node-level stage recognition from holistic appearance, without explicit modeling of localized states.

All experiments were conducted within a unified PyTorch framework on a server equipped with eight NVIDIA GeForce RTX 3090 GPUs (24 GB VRAM each). All reported baselines were trained solely on GSO-Net without supplementary industrial data. For YOLO-based detection models, training was performed for 100 epochs with input resolution $640 \times 640$. Other state-of-the-art detection models were trained and evaluated using the default resolutions and hyperparameters specified in their official codebases. For classification networks, input resolutions varied according to the backbone, as detailed in Table~\ref{tab:classification_results}. Unless otherwise specified, all methods followed their official implementations and default configurations.

\subsection{Task 1: Joint Detection of Microscopic States and Macroscopic Steps}
\label{sec:task1}

Task 1 is the main benchmark task of GSO-Net. It tests whether a detector can recover procedural meaning from sparse local evidence under a unified step-state formulation. The challenge is twofold. First, microscopic states and macroscopic steps differ sharply in spatial scale: decisive evidence is often confined to tiny contact regions, whereas the step target spans the full frame. Second, stage recognition depends on localized functional cues rather than only global scene layout. The benchmark therefore requires both fine-grained evidence grounding and robust scene-level stage discrimination.

\begin{figure}[!t]
\centering
\includegraphics[width=\columnwidth]{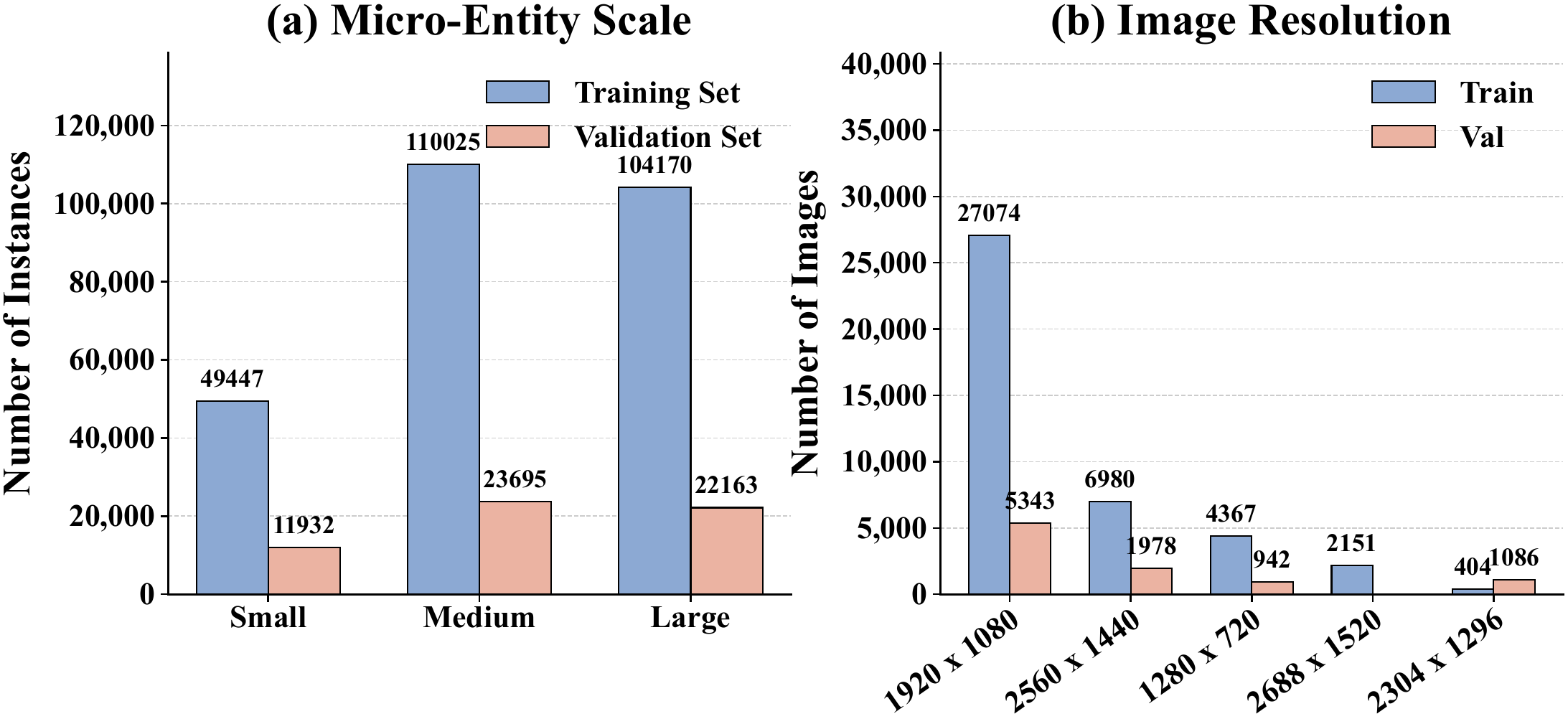}
\caption{Scale disparity and input diversity in GSO-Net. \textbf{Left:} Bounding-box area distribution under the COCO size protocol. Safety-critical operational cues are heavily concentrated in the extremely small-object regime. \textbf{Right:} Resolution distribution of the raw surveillance frames. The benchmark combines multi-megapixel inputs with tiny state indicators, which makes precise localization difficult under practical resizing constraints.}
\label{fig:dataset_joint_distribution}
\end{figure}

Fig.~\ref{fig:dataset_joint_distribution} summarizes this difficulty. The raw surveillance frames are high-resolution and visually diverse. However, many states that determine stage recognition are encoded by only a few local pixels. In particular, clamp status and hose connection status are often much smaller than the main foreground actors. As a result, the benchmark is not a standard object detection problem. It is a multi-scale state-grounding problem with strong semantic asymmetry.

\subsubsection{Lightweight Detection Baselines}

We first evaluate lightweight detectors that are relevant to edge deployment.

\begin{table}[!t]
\caption{Task 1 results on lightweight detectors (\%).}
\label{tab:yolo_detection}
\centering
\setlength{\tabcolsep}{3pt}
\resizebox{\columnwidth}{!}{
\begin{tabular}{l c c c c c c c c}
\toprule
\textbf{Model} & \textbf{Size} & \textbf{Ep.} & \textbf{Params(M)} & \textbf{AP$_{50}$} & \textbf{AP$_{50-95}$} & \textbf{AP$_{S}$} & \textbf{AP$_{M}$} & \textbf{AP$_{L}$} \\
\midrule
YOLOv8-n \cite{ref_yolov8}   & 640 & 100 & 3.01          & 42.1          & 35.0          & 3.2          & 11.6          & 33.6 \\
YOLOv9-t \cite{ref_yolov9}   & 640 & 100 & \textbf{1.98} & \textbf{44.0} & \textbf{36.5} & 3.0          & 13.1          & 33.5 \\
YOLOv10-n \cite{ref_yolov10} & 640 & 100 & 2.27          & 37.5          & 31.4          & 3.5          & 12.5          & 38.0 \\
YOLO11-n \cite{ref_yolov11}  & 640 & 100 & 2.60          & 41.0          & 33.0          & \textbf{4.9} & 12.2          & 30.1 \\
YOLO26-n \cite{ref_yolov26}  & 640 & 100 & 2.38          & 41.7          & 34.5          & 4.1          & \textbf{13.9} & \textbf{42.2} \\
\bottomrule
\end{tabular}
}
\end{table}

The results are reported in Table~\ref{tab:yolo_detection}. YOLOv9-t provides the strongest overall trade-off among the lightweight models. It achieves $44.0\%$ $\mathrm{AP}_{50}$ and $36.5\%$ $\mathrm{AP}_{50\text{-}95}$ with only $1.98$M parameters. YOLO26-n gives the highest $\mathrm{AP}_{M}$ and $\mathrm{AP}_{L}$, at $13.9\%$ and $42.2\%$, respectively.

The main bottleneck is exposed by $\mathrm{AP}_{S}$. Across all five models, $\mathrm{AP}_{S}$ remains between $3.0\%$ and $4.9\%$. The gap between $\mathrm{AP}_{S}$ and $\mathrm{AP}_{L}$ is large for every model. These numbers show that the critical difficulty of GSO-Net is not generic object presence. It is the recovery of small, state-sensitive evidence after aggressive feature downsampling.

A second observation is that parameter efficiency does not translate into reliable SOP understanding. The best lightweight baseline identifies major foreground actors and some stable states. Yet the most safety-critical microscopic evidence is still missed. These models should therefore be interpreted as deployment baselines rather than complete solutions.

\subsubsection{Comparison with Stronger Detectors}

We next evaluate stronger transformer-based and open-vocabulary detectors.

\begin{table}[!t]
\caption{Task 1 results on stronger detectors (\%).}
\label{tab:sota_detection}
\centering
\setlength{\tabcolsep}{4pt}
\resizebox{\columnwidth}{!}{
\begin{tabular}{l c c c c c c}
\toprule
\textbf{Method} & \textbf{Venue} & \textbf{AP$_{50}$} & \textbf{AP$_{50-95}$} & \textbf{AP$_{S}$} & \textbf{AP$_{M}$} & \textbf{AP$_{L}$} \\
\midrule
RT-DETR \cite{ref_rtdetr}                      & CVPR'24  & 35.1 & 28.7 & 4.0  & 11.2 & 32.4 \\
YOLO-World \cite{ref_yoloworld}                & CVPR'24  & 45.6 & 36.3 & 6.8  & 15.7 & 35.0 \\
Relation-DETR-ResNet50 \cite{ref_relationdetr} & ECCV'24  & 43.8 & 35.1 & 9.8  & 19.8 & 42.9 \\
Relation-DETR-SwinL \cite{ref_relationdetr}    & ECCV'24  & \textbf{56.4} & \textbf{46.3} & \textbf{13.3} & \textbf{23.4} & \textbf{55.1} \\
\midrule
Hyper-YOLO \cite{ref_hyperyolo}                & TPAMI'25 & 40.6 & 34.1 & 3.4  & 11.2 & 33.4 \\
Mr.DETR-ResNet50 \cite{ref_mrdetr}             & CVPR'25  & 45.0 & 37.5 & 11.9 & 19.6 & 46.7 \\
Mr.DETR-SwinL \cite{ref_mrdetr}                & CVPR'25  & 46.1 & 37.2 & 12.9 & 20.8 & 45.2 \\
\midrule
RF-DETR \cite{ref_rfdetr}                      & ICLR'26 & 37.5 & 30.3 & 1.6  & 10.8 & 38.9 \\
\bottomrule
\end{tabular}
}
\end{table}

Table~\ref{tab:sota_detection} shows that stronger detectors yield clear gains. Relation-DETR-SwinL provides the best overall results on all reported metrics. It reaches $56.4\%$ $\mathrm{AP}_{50}$ and $46.3\%$ $\mathrm{AP}_{50\text{-}95}$. It also gives the best scale-aware performance, including $13.3\%$ $\mathrm{AP}_{S}$, $23.4\%$ $\mathrm{AP}_{M}$, and $55.1\%$ $\mathrm{AP}_{L}$.

Nevertheless, the small-object problem is not removed. Even for Relation-DETR-SwinL, $\mathrm{AP}_{S}$ remains far below $\mathrm{AP}_{L}$. Stronger global attention therefore helps, but decisive contact-level evidence remains far from solved.

YOLO-World provides an informative reference point. It reaches $45.6\%$ $\mathrm{AP}_{50}$ and improves $\mathrm{AP}_{S}$ to $6.8\%$, which is higher than all lightweight closed-set YOLO baselines. This suggests that language-aligned pre-training is helpful for broad semantic transfer. However, the gain remains limited on the smallest and most mechanically defined categories. The benchmark requires more than semantic name transfer. It requires precise state grounding.

\subsubsection{Category-wise Diagnosis}

The overall metrics in Tables~\ref{tab:yolo_detection} and \ref{tab:sota_detection} are informative, but they do not fully reveal which categories remain unsolved. We therefore further report the per-class $\mathrm{AP}_{50}$ results in Table~\ref{tab:per_class_ap}.

\begin{table*}[!t]
\caption{Comprehensive per-class Average Precision ($\text{AP}_{50}$, \%) for all 24 categories across selected models. Best results are \textbf{bolded}, and second-best are \underline{underlined}.}
\label{tab:per_class_ap}
\centering
\renewcommand{\arraystretch}{1.1}
\setlength{\tabcolsep}{5pt}
\resizebox{\textwidth}{!}{
\begin{tabular}{l c c | c c c c | c c c | c}
\toprule
\textbf{Category} & \textbf{YOLOv9-t} & \textbf{YOLO26-n} & \textbf{RT-DETR} & \textbf{YOLO-World} & \textbf{Relation-DETR-R50} & \textbf{Relation-DETR-SwinL} & \textbf{Hyper-YOLO} & \textbf{Mr.DETR-R50} & \textbf{Mr.DETR-SwinL} & \textbf{RF-DETR} \\
\midrule
\multicolumn{11}{l}{\textit{Microscopic Entity States ($E_{micro}$)}} \\
\midrule
00: tanker\_moving\_true      &  0.0 &  0.0 &  0.0 &  0.0 &  0.0 &  0.0 &  0.0 &  0.0 &  0.0 &  0.0 \\
01: tanker\_moving\_false     & 73.4 & 74.9 & 58.7 & 80.1 & \underline{88.3} & \textbf{89.5} & 58.2 & 65.7 & 68.3 & 61.6 \\
02: tanker\_stopped\_true     & 93.7 & 92.8 & 78.2 & 92.6 & \underline{94.7} & \textbf{99.3} & 85.7 & 90.6 & 88.6 & 92.9 \\
03: tanker\_stopped\_false    & 81.5 & 71.6 & 76.8 & \underline{87.1} & 83.3 & \textbf{95.0} & 76.8 & 83.0 & 81.5 & 67.2 \\
04: staff\_standing           & 55.6 & 50.5 & 38.5 & 60.4 & \underline{62.9} & \textbf{67.2} & 53.5 & 36.0 & 37.3 & 53.8 \\
05: staff\_bending            & 38.5 & \underline{62.1} & 51.1 & 34.4 & 30.5 & \textbf{64.0} & 10.0 & 17.6 & 37.5 & 30.1 \\
06: staff\_walking            & 32.2 & 28.9 & 28.2 & 38.7 & \underline{41.4} & \textbf{48.2} & 30.1 & 31.5 & 27.8 & 34.8 \\
07: staff\_operating          & 56.3 & 52.5 & 45.0 & 62.3 & \textbf{68.2} & \underline{66.2} & 56.6 & 42.5 & 41.0 & 51.8 \\
08: clamp\_connected          &  0.1 &  0.5 &  0.5 &  1.3 &  \underline{2.9} & \textbf{5.3} &  0.3 &  0.5 &  1.3 &  0.4 \\
09: clamp\_disconnected       &  0.0 &  0.0 &  0.0 &  0.0 &  0.0 &  0.0 &  0.0 &  0.0 &  0.0 &  0.0 \\
10: clamp\_floating           &  0.0 &  0.0 &  0.0 &  0.1 &  \underline{3.2} & \textbf{6.0} &  0.0 &  0.2 &  0.1 &  0.0 \\
11: extinguisher\_deployed    & 73.6 & 70.8 & 53.3 & \underline{81.6} & 78.2 & \textbf{83.3} & 73.9 & 52.7 & 54.0 & 63.6 \\
12: extinguisher\_carried     & 23.4 & 15.3 & 23.1 & \underline{45.3} & 41.4 & \textbf{70.7} & 19.9 & 19.4 & 19.1 & 21.7 \\
13: discharge\_connected      & 34.0 & 35.2 & \underline{48.4} & 45.9 & \textbf{56.2} & 47.8 & 34.1 & 14.0 & 15.6 & 20.1 \\
14: discharge\_disconnected   &  2.0 &  1.1 &  1.3 &  \underline{2.8} &  2.3 & \textbf{15.7} &  1.6 &  1.2 &  2.6 &  2.0 \\
\midrule
\multicolumn{11}{l}{\textit{Macroscopic Procedural Steps ($S_{macro}$)}} \\
\midrule
15: Step0\_Idle               & 87.0 & 84.2 & 81.4 & 83.0 & 78.7 & 87.4 & \textbf{93.6} & \underline{90.2} & 88.7 & 87.1 \\
16: Step1\_Entry              & 54.0 & 58.9 & 53.1 & \underline{62.8} & 53.0 & \textbf{73.9} & 57.2 & 50.5 & 48.3 & 58.8 \\
17: Step2\_Safety\_Protection & 41.0 & 32.7 &  6.5 & \underline{47.2} & 16.1 & \textbf{68.7} & 46.4 & 36.0 & 38.5 & 19.3 \\
18: Step3\_Safety\_Check      & 28.5 & 26.3 &  5.3 & \underline{29.4} & 13.2 & \textbf{50.9} & 28.3 & 18.7 & 14.9 & 18.9 \\
19: Step4\_Pre\_Unload\_Confirm& \underline{37.0} & 22.7 & 25.1 & 23.7 & 25.1 & \textbf{47.2} & 28.2 & 28.2 & 28.9 & 10.5 \\
20: Step5\_Oil\_Unloading     & \textbf{56.1} & \underline{52.6} & 35.4 & 42.2 & 43.1 & 41.1 & 48.6 & 51.8 & 41.8 & 39.8 \\
21: Step6\_Supervision        & 87.6 & 81.6 & 68.8 & 87.4 & 90.6 & \textbf{96.2} & 82.7 & \underline{95.3} & 86.5 & 88.9 \\
22: Step7\_Receipt\_Acceptance& \underline{29.6} & 27.0 & 18.9 & 21.2 & 21.0 & \textbf{40.7} & 28.0 & 20.0 & 16.2 & 18.4 \\
23: Step8\_Post\_Processing   & \underline{25.7} & 17.1 &  8.8 & 19.5 & 12.7 & \textbf{33.9} & 19.1 & 16.2 & 17.1 & 21.7 \\
\bottomrule
\end{tabular}
}
\end{table*}

Several clear patterns emerge.

\textbf{(1) Frequent and visually salient states are learned well.}  
Categories such as \textit{tanker\_stopped\_true}, \textit{staff\_standing}, \textit{staff\_operating}, and \textit{extinguisher\_deployed} achieve relatively high scores. These classes are more frequent in the dataset and supported by larger visual extent.

\textbf{(2) Mechanically decisive states remain difficult.}  
The main failure occurs on contact-level categories. \textit{clamp\_connected} remains at or below $5.3\%$ across all models. \textit{clamp\_floating} remains at or below $6.0\%$. \textit{discharge\_disconnected} reaches $15.7\%$ only for Relation-DETR-SwinL and stays near zero for most other methods. \textit{clamp\_disconnected} is essentially unresolved.

\textbf{(3) Stable workflow phases are easier than transient phases.}  
Among the macroscopic steps, \textit{Step0\_Idle} and \textit{Step6\_Supervision} are reliably detected by several models. By contrast, the transition phases \textit{Step2}, \textit{Step3}, \textit{Step4}, \textit{Step5}, and \textit{Step7} remain lower across most models. These steps differ by localized evidence rather than overall scene layout.

\textbf{(4) Strong object detection does not guarantee correct SOP understanding.}  
A model may detect the tanker, the worker, and the extinguisher, yet still miss the decisive connection state or the correct procedural phase. The unresolved categories are not generic object classes. They are rare state transitions, small contact points, and short procedural stages.

\subsection{Task 2: Frame-Level SOP Step Classification}
\label{sec:task2}

Task 2 is introduced as a diagnostic reference rather than a parallel core benchmark target. It isolates frame-level macroscopic step recognition by removing explicit box supervision and localized state grounding. Each classifier receives a full frame and predicts one of the 9 SOP steps from holistic visual evidence alone. We evaluate two YOLO classification baselines, YOLOv8-cls and YOLO26-cls, together with recent generic backbones, including ConvNeXt-V2 \cite{ref_convnextv2}, EVA-02 \cite{ref_eva02}, and MambaOut \cite{ref_mambaout}. Although the output space is simpler than in Task 1, accurate step prediction remains difficult because success still depends on sparse local cues and on whether the model captures the operational meaning of visible objects rather than only dominant scene content.

\begin{figure}[!t]
\centering
\includegraphics[width=\columnwidth]{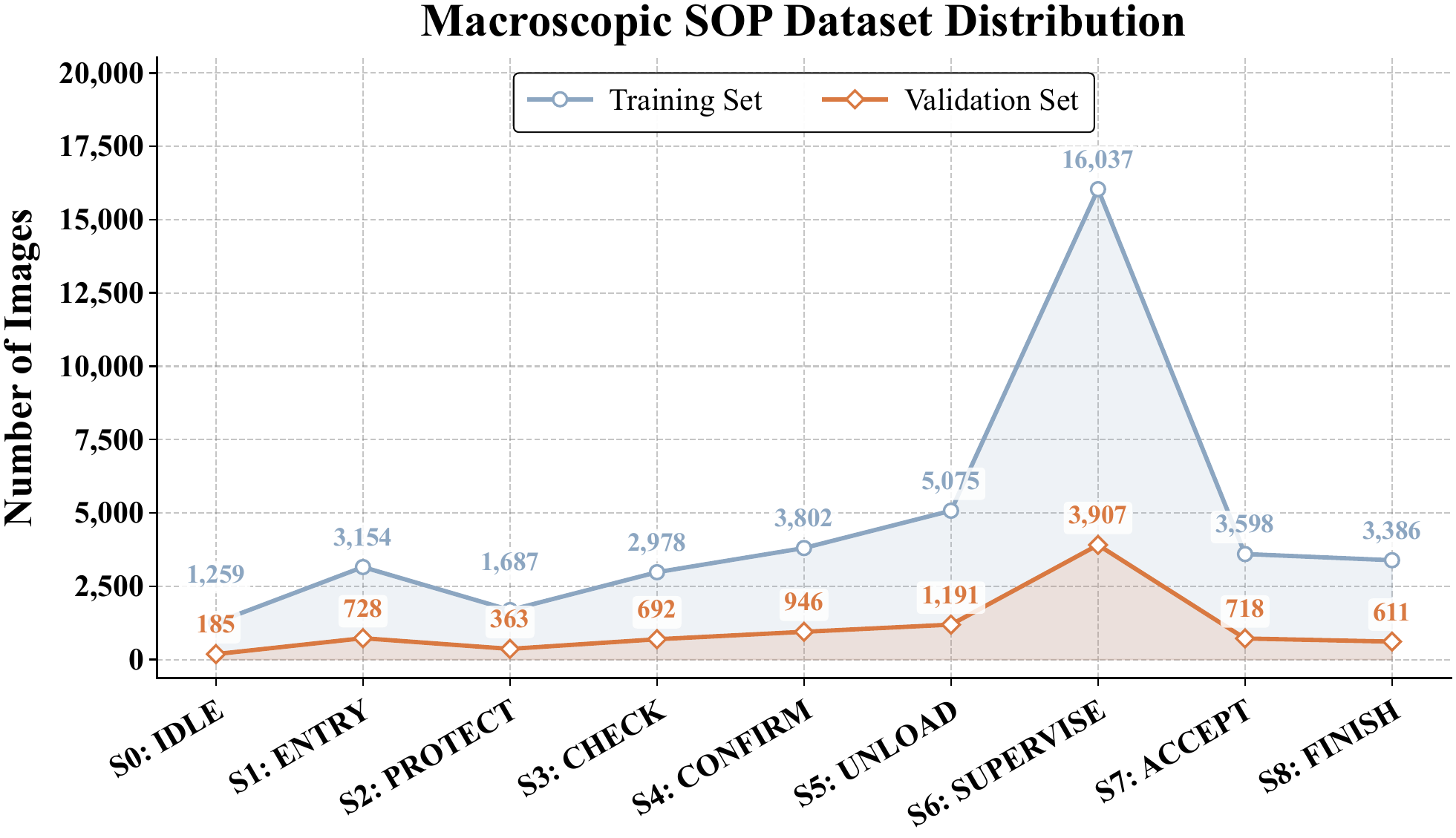}
\caption{Frame distribution of the 9 macroscopic SOP steps in the training and validation sets. Long-duration phases dominate the benchmark, whereas several transitional steps remain comparatively rare.}
\label{fig:cls_dataset_dist}
\end{figure}

Fig.~\ref{fig:cls_dataset_dist} illustrates the temporal imbalance in the benchmark. Long-duration phases dominate frame counts, whereas short transitional stages remain sparse. Top-1 accuracy is therefore insufficient on its own, so Macro F1 is reported alongside Top-1 and Top-5. The task tests whether SOP recognition can be reasonably approximated by scene-level classification before localized state evidence is introduced.

\begin{table}[!t]
\caption{Frame-level macroscopic SOP step classification results (\%).}
\label{tab:classification_results}
\centering
\setlength{\tabcolsep}{4pt}
\resizebox{\columnwidth}{!}{
\begin{tabular}{l c c c c c}
\toprule
\textbf{Model} & \textbf{Size} & \textbf{Epochs} & \textbf{Top-1} & \textbf{Top-5} & \textbf{Macro F1} \\
\midrule
YOLOv8-cls      & 640 & 100 & 41.7 & 83.8 & 8.9  \\
YOLO26-cls      & 640 & 100 & \textbf{44.3} & \textbf{86.2} & \textbf{16.1} \\
ConvNeXt-V2     & 224 & 100 & 41.8 & 75.3 & 6.8  \\
EVA-02          & 224 & 100 & 43.6 & 80.9 & 14.1 \\
MambaOut        & 224 & 100 & 41.8 & 78.9 & 6.6  \\
\bottomrule
\end{tabular}
}
\end{table}

Table~\ref{tab:classification_results} shows that all models fall within a narrow Top-1 range ($41.7\%-44.3\%$). YOLO26-cls provides the strongest overall classification result, reaching $44.3\%$ Top-1, $86.2\%$ Top-5, and $16.1\%$ Macro F1, but the absolute level remains limited. The large Top-1/Top-5 gap indicates that coarse procedural context is captured, but precise stage boundaries remain unresolved. Short transition phases are especially difficult because their decisive cues are sparse and localized.

These results indicate that frame-level step classification remains under-constrained when localized state supervision is absent. In the sparse polling setting, holistic appearance alone cannot reliably capture subtle contact-level cues and short transitional evidence that distinguish adjacent SOP stages. As a result, image-level classifiers may recognize dominant or visually stable phases, but they remain unreliable for fine-grained stage discrimination across the workflow.

The limitation is structural: many step boundaries are defined by \textit{functional state transitions} rather than global scene changes. Holistic classifiers can learn broad stage priors, but remain weak when decisive signals are sparse and localized. The issue is therefore not only temporal ambiguity, but also the model’s ability to capture the operational meaning of small objects, contact relations, and state-bearing components.

Task 2 thus serves as a \textit{diagnostic reference task}. It shows that SOP recognition cannot be reduced to holistic image classification, because the required temporal logic is expressed through localized functional state transitions rather than coarse scene appearance. This directly motivates Task 1, where macroscopic steps provide procedural context and microscopic states supply the local evidence needed for robust stage recognition. From an engineering perspective, GSO-Net should therefore be read as a benchmark for practical visual state machines under sparse polling, small-object difficulty, and long-tailed operational evidence.

\subsection{Qualitative Analysis and Error Statistics}
\label{sec:error_analysis}

\begin{figure*}[!t]
\centering
\includegraphics[width=1\textwidth]{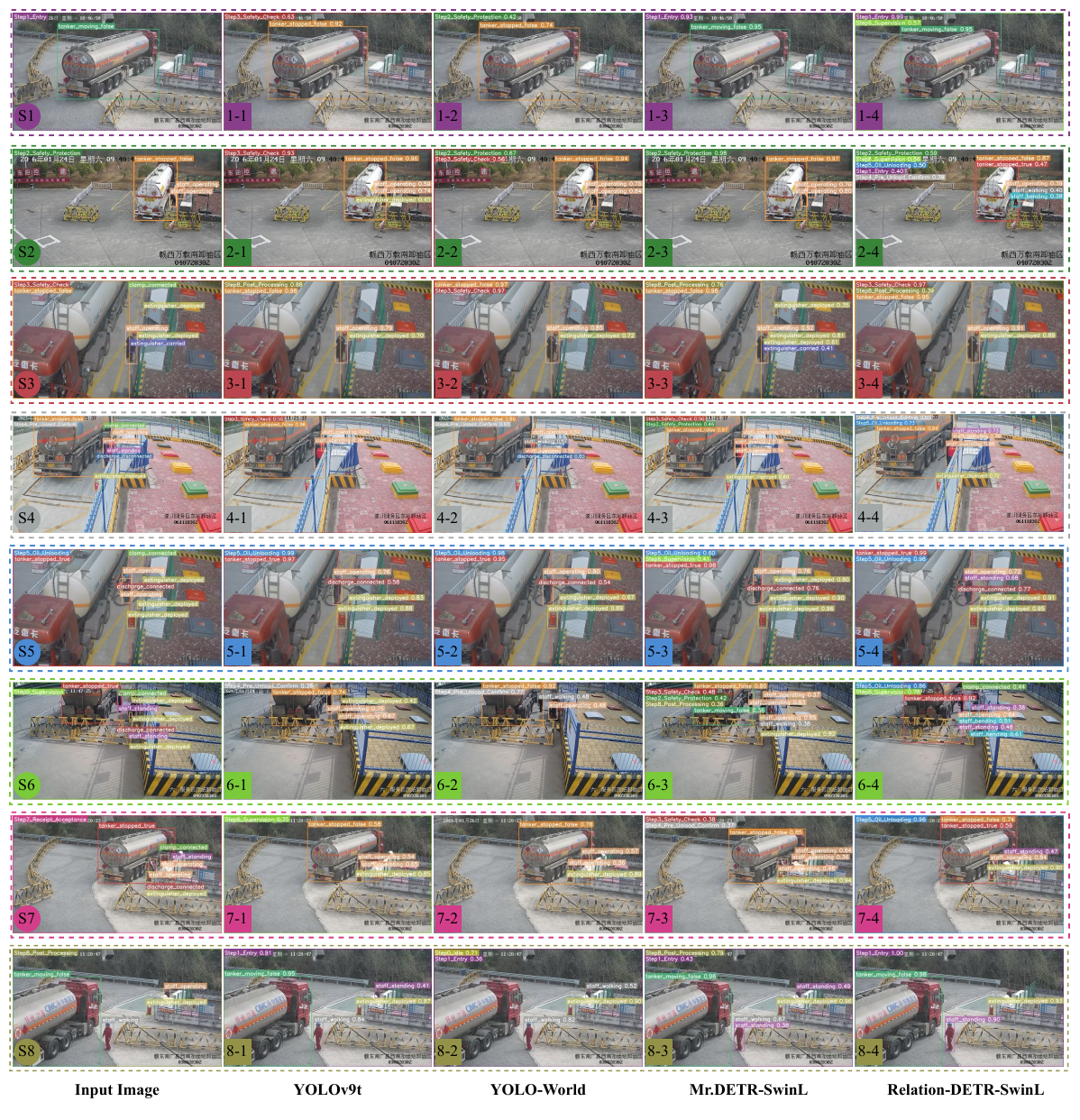}
\caption{Qualitative comparison of representative detectors across SOP phases $S_1$--$S_8$. \textbf{Column 1:} Input images with ground-truth phase labels. \textbf{Columns 2--5:} Predictions from YOLOv9-t, YOLO-World, Mr.DETR-SwinL, and Relation-DETR-SwinL. Three recurrent failure modes are observed: adjacent-step confusion, omission of critical micro-states, and stage predictions inconsistent with visible evidence.}
\label{fig:qualitative_analysis}
\end{figure*}

To further examine the failure modes behind the quantitative results, Fig.~\ref{fig:qualitative_analysis} provides representative predictions from YOLOv9-t, YOLO-World, Mr.DETR-SwinL, and Relation-DETR-SwinL. All predictions are generated with a confidence threshold of $0.35$ and an IoU threshold of $0.40$. Row--column indexing is used for reference, where the row denotes the SOP phase ($S_1$--$S_8$) and the column denotes the detector output.

Three recurrent behaviors are observed.

\textbf{(1) Limited correct phase grounding.}  
Accurate and visually consistent phase prediction appears only in a small subset of cells, such as 1-3, 2-3, 4-2, 5-1, 5-2, and 5-4. When decisive cues are clearly visible, current detectors can recover the correct phase.

\textbf{(2) Phase ambiguity.}  
Many cells, including 1-4, 2-2, 2-4, 3-4, 4-4, 5-3, 6-4, and 8-3, contain overlapping or poorly ranked predictions. The ground-truth phase is often present, but not ranked first. This behavior is consistent with the Top-1/Top-5 gap observed in Task 2.

\textbf{(3) Semantic inconsistency.}  
Cases such as 4-3, 6-3, and 7-3 show predictions that contradict visible operational evidence.

These cases indicate three main error sources: spatiotemporal ambiguity between adjacent phases, omission of decisive microscopic evidence such as clamp or hose states, and compositional inconsistency between predicted stages and visible cues. Among the four detectors, Relation-DETR-SwinL usually provides the most complete visual evidence and the strongest phase sensitivity, consistent with Table~\ref{tab:sota_detection}. However, even the strongest detector still suffers from ranking ambiguity and incomplete evidence support.

\subsection{Summary of Empirical Findings}
\label{sec:exp_summary}

The experiments support four main conclusions. First, the strict cross-node split makes generalization to unseen logistics nodes difficult. Second, current detectors recognize major actors and stable phases, but still fail on the microscopic contact states that determine procedural status. Third, image-level step classification is biased toward dominant phases and does not recover short transitions reliably. Fourth, the remaining gap is structural rather than a simple matter of stronger backbones.

Overall, the benchmark results suggest that accurate SOP monitoring cannot be achieved by scene-level recognition alone. Fine-grained state grounding is essential, which is precisely why Task 1, rather than Task 2, constitutes the core benchmark evaluation of GSO-Net.

\section{Dataset Advantages}
\label{sec:dataset_advantages}

The experimental results indicate that GSO-Net is neither a standard object detection benchmark nor a conventional frame-classification benchmark. Performance is reasonable on salient objects and dominant workflow phases, but drops sharply when the decision depends on tiny contact states, rare events, or short transitions. This section discusses the implications of these findings for vision-based safety monitoring at hazardous freight logistics nodes.

\textbf{From Object Detection to Stage-Aware Node Understanding.}
Mainstream detectors are designed to answer what is present and where it is \cite{fasterrcnn,fpn,detr,dino,glip,ref_groundingdino,ref_yoloworld,ref_owovd}. GSO-Net requires an additional capability: linking localized visual evidence to procedural stage understanding at an infrastructure-side transportation node.

This difference appears in the results. In Table~\ref{tab:per_class_ap}, visually salient categories such as tanker and personnel states achieve high scores, whereas decisive categories such as \textit{clamp\_connected}, \textit{clamp\_floating}, and \textit{discharge\_disconnected} remain difficult. A similar pattern appears for macroscopic steps, where stable phases are recognized more reliably than short transitions. Correct detection therefore does not guarantee correct SOP understanding.

\textbf{Why Sparse Infrastructure Polling Makes the Problem Hard.}
Many existing video models and procedural benchmarks assume dense temporal continuity \cite{timesformer,videomae,assembly101,epic100,ego4d,openpack}. GSO-Net follows a different setting. In infrastructure-side deployment, cameras are observed through sparse round-robin polling, so the visual stream is discontinuous and transitions may appear in only one frame.

This setting explains the weakness of frame-level classification in Task 2. In Table~\ref{tab:classification_results}, the accuracy of Top-1 remains limited, while Macro F1 is substantially lower. Short transition phases are especially difficult because sparse observation shifts stage discrimination toward localized state transitions rather than global scene changes. Pure frame-level classification is therefore insufficient for practical understanding of SOP.

\textbf{Toward State-Aware SOP Reasoning under Sparse Polling.}
The benchmark results suggest that stronger backbones alone are unlikely to close the gap. Larger detectors improve overall metrics, but the hardest microscopic states, rare cases, and transient phases remain unresolved. These findings indicate that future progress will depend not only on stronger perception, but also on mechanisms that preserve local state evidence and relate it to sparse procedural context.

A practical direction for sparse visual polling may include:
\begin{enumerate}
    \item \textbf{Micro-Perception Module:} to preserve state-sensitive evidence such as contact points.
    \item \textbf{Sparse Temporal Memory:} to accumulate partial observations across discontinuous frames.
    \item \textbf{Stage-Aware Reasoning Module:} to relate observed states to broader workflow context.
\end{enumerate}
This decomposition is consistent with the benchmark findings: the first component targets microscopic-state failure, the second addresses sparse polling, and the third addresses ambiguous stage prediction.

\section{Discussion}
\label{sec:discussion}

1) \textbf{Dataset Challenges.} GSO-Net poses several challenges. \textbf{Hierarchical Step-State Coupling:} Unlike conventional detection benchmarks, it requires joint understanding of macroscopic SOP stages and microscopic operational states, so object presence alone is insufficient. \textbf{Sparse and Subtle Evidence:} Under round-robin polling, stage inference must rely on isolated frames and highly localized cues, such as clamp status and hose-port connection, which are often extremely small and visually subtle. \textbf{Real-World Difficulty:} The benchmark further exhibits severe long-tail and scale imbalance, together with occlusion, low illumination, adverse weather, and cross-node viewpoint variation.

2) \textbf{Limitations.} This study remains limited in three main aspects. \textbf{Domain Scope:} The current benchmark is restricted to petrochemical unloading scenarios at logistics nodes. \textbf{Temporal Supervision:} Its independently sampled frames do not yet support richer temporal modeling or explicit association across sparse observations. \textbf{Reasoning Scope:} Although GSO-Net provides hierarchical step-state annotations and systematic baselines, it does not yet implement an explicit rule-based or consistency-aware reasoning protocol for SOP verification.

3) \textbf{Future Work.}
Future work will extend the benchmark to more hazardous-freight and infrastructure-side scenarios, introduce sparse temporal annotations and association signals, and develop stage-aware models that use the hierarchical relationship between microscopic states and macroscopic steps. More broadly, this direction may improve safety-oriented visual round-robin inspection in intelligent transportation and infrastructure monitoring, while advancing visual detection from perception toward structured semantic and procedure-aware understanding. Cross-node continual evaluation will also be important for reducing the gap between benchmark performance and real-world reliability.

\section{Conclusion}
\label{sec:conclusion}

This paper introduced GSO-Net, a large-scale benchmark for visual SOP understanding in petrochemical hazardous freight-transfer scenarios under sparse infrastructure-side polling. By coupling 9 macroscopic procedural steps with 15 microscopic operational states, GSO-Net formulates hazardous freight-transfer monitoring as a hierarchical stage-aware state-grounding problem. Experiments on 64 real petrochemical logistics nodes show that current models remain effective on salient objects and dominant phases, but still struggle with contact-level state grounding, transient phase recognition, and stage-consistent understanding under sparse and localized evidence. These findings show that reliable visual safety monitoring requires moving beyond surface-level detection toward structured procedural understanding. GSO-Net provides a practical benchmark for future research on stage-aware reasoning and safety-oriented visual inspection in intelligent transportation and freight logistics.

\appendices

\section{Hyperparameter Configurations}
To improve reproducibility, the YOLO-series baselines were trained under a unified training configuration in the Ultralytics framework. All YOLO-based models were trained for 100 epochs with a batch size of 16 and 8 dataloader workers. Early stopping with a patience of 50 epochs was applied to reduce overfitting on the imbalanced validation set.

Optimization was performed using stochastic gradient descent (SGD). The initial learning rate was set to $0.01$, the final learning-rate ratio was set to $0.01$, and the momentum was set to $0.937$. A warmup schedule was used during the first 3.0 epochs, with the warmup momentum initialized at $0.8$.

For fair comparison, the input resolution was fixed at $640 \times 640$ for all YOLO-based detection and classification baselines. Standard data augmentation was applied during training, including Mosaic augmentation, random horizontal flipping, color jitter, and spatial scaling. Mosaic augmentation was disabled in the final 10 epochs to improve fine-grained localization stability. Validation was performed at the same input resolution and without data augmentation.

For non-YOLO state-of-the-art baselines, we followed the official implementations and default training protocols released by the original authors, including their recommended input resolutions and optimization settings, unless otherwise specified. In this way, each method was evaluated under a configuration consistent with its native design, while the YOLO-series baselines were compared under a unified experimental protocol.

\section{Annotation Guidelines and Tooling}
Manual annotation was conducted using the open-source CVAT platform (v2.4). Annotators followed a visibility-first principle: operational states were assigned only when the image provided sufficient visual evidence.

This rule was especially important for mechanically sensitive states. For example, if the locking condition of a grounding clamp was occluded by the operator's hand, body, or another object, the instance was labeled as ``floating'' rather than ``connected.'' Similarly, hose-related states were assigned only when the connection or disconnection status was visually verifiable. For engaged states such as \textit{clamp\_connected}, the bounding box was required to tightly enclose the functional contact point rather than the surrounding object region.

When visual evidence was weak because of occlusion, low illumination, motion blur, or viewpoint limitation, annotators were instructed to avoid speculative assignment and retain ambiguous cases for subsequent review.

\section*{Acknowledgment}
This work was supported by the National Natural Science Foundation of China under Grants 62473187 and 62401244, and by the Jiangxi Province Key Research and Development Program under Project No. 20261BCE310029.

\ifCLASSOPTIONcaptionsoff
  \newpage
\fi

\end{document}